\documentclass[11pt, a4paper, oneside, reqno]{amsart}
\usepackage{comment}
\usepackage[usenames, dvipsnames]{color}
\definecolor{darkblue}{rgb}{0.0, 0.0, 0.45}
\definecolor{lightblue}{RGB}{240,248,255}
\definecolor{lightblue2}{rgb}{0.68, 0.85, 0.9}
\definecolor{lightcyan}{rgb}{0.88, 1.0, 1.0}
\definecolor{palepink}{rgb}{0.98, 0.85, 0.87}

\usepackage[colorlinks	= true,
raiselinks	= true,
linkcolor	= darkblue, 
citecolor	= Mahogany,
urlcolor	= ForestGreen,
pdfauthor	= {Peyman Mohajerin Esfahani},
pdftitle	= {},
pdfkeywords	= {},
pdfsubject	= {},
plainpages	= false]{hyperref}

\usepackage{dsfont,amssymb,amsmath,enumitem} 
\usepackage{amsfonts,dsfont,mathtools, mathrsfs,amsthm} 
\usepackage[amssymb, thickqspace]{SIunits}
\usepackage{fancyhdr,mdframed,nicefrac}

\usepackage{epsfig}
\usepackage{graphicx}
\usepackage{float}
\usepackage{caption}
\usepackage{subcaption}

\usepackage{courier}

\usepackage{multirow}
\usepackage{bigstrut}

\allowdisplaybreaks
\date{\today}
\makeatletter
\def\@settitle{\begin{center}%
		\baselineskip14\p@\relax
		\normalfont\LARGE\scshape\bfseries
		\@title
	\end{center}%
}

\def\@setauthors{%
  \begingroup
  \def\thanks{\protect\thanks@warning}%
  \trivlist
  \centering\footnotesize \@topsep30\p@\relax
  \advance\@topsep by -\baselineskip
  \item\relax
  \author@andify\authors
  \def\\{\protect\linebreak}%
  \authors%
  \ifx\@empty\contribs
  \else
    ,\penalty-3 \space \@setcontribs
    \@closetoccontribs
  \fi
  \endtrivlist
  \endgroup
}

\makeatother
\makeatletter

\def\subsection{\@startsection{subsection}{2}%
	\z@{.5\linespacing\@plus.7\linespacing}{.5\linespacing}%
	{\normalfont\large\bfseries}}

\def\subsubsection{\@startsection{subsubsection}{3}%
	\z@{.5\linespacing\@plus.7\linespacing}{.5\linespacing}%
	{\normalfont\itshape}}

\usepackage{multirow}
\usepackage{bigstrut}
\usepackage{wrapfig}
\usepackage{caption}

\renewcommand{\geq}{\geqslant}

\renewcommand{\leq}{\leqslant}

\DeclareSymbolFont{symbolsC}{U}{pxsyc}{m}{n}

\DeclareMathOperator*{\argmin}{argmin}

\usepackage{bm}
\usepackage{amsthm}

\usepackage{lineno,hyperref}
\modulolinenumbers[5]
\usepackage{color}
\usepackage{afterpage}
\usepackage{floatpag}
\usepackage{amsmath,bm,amssymb}
\usepackage{comment}
\usepackage{xcolor}
\usepackage{mathtools}
 \usepackage{subcaption}
\definecolor{referee1}{rgb}{1.0, 0.25, 0.25}
\definecolor{referee2}{rgb}{0.0, 0.44, 1.0}
\definecolor{myred}{rgb}{1.0, 0.0, 0.0}

\definecolor{ms1}{rgb}{0, 0, 0}
\definecolor{ms2}{rgb}{0, 0, 0}
\definecolor{hg1}{rgb}{0, 0, 0}
\definecolor{hg2}{rgb}{0, 0, 0}

\definecolor{ms1_rv1}{rgb}{0.0, 0.0, 0.0}

\usepackage{algorithm}
\usepackage{algpseudocode}
\usepackage{hyperref} 

\title[Gaussian Process Regression for Maximum Entropy Distribution]{
Gaussian Process Regression for Maximum Entropy Distribution
}
 \thanks{Corresponding author: Mohsen Sadr}
 \thanks {Email: sadr@mathcces.rwth-aachen.de}
\thanks{Mohsen Sadr and Manuel Torrilhon: MATHCCES, Department of Mathematics, RWTH Aachen University, Schinkestrasse 2, D-52062 Aachen,
Germany. M. Hossein Gorji: MCSS, Ecole Polytechnique F{\'e}d{\'e}rale de Lausanne (EPFL), CH-1015 Lausanne, Switzerland.}
\date{October 1, 2020}

\begin{document}

\begin{abstract}
Maximum-Entropy Distributions offer an attractive family of probability densities suitable for moment closure problems. Yet finding the Lagrange multipliers which parametrize these distributions, turns out to be a computational bottleneck for practical closure settings. Motivated by recent success of Gaussian processes, we investigate the suitability of Gaussian priors to approximate the Lagrange multipliers as a map of a given set of moments. Examining various kernel functions, the hyperparameters are optimized by maximizing the log-likelihood. The performance of the devised data-driven Maximum-Entropy closure is studied for couple of test cases including relaxation of non-equilibrium distributions governed by Bhatnagar-Gross-Krook and Boltzmann kinetic equations.
\end{abstract}

\maketitle

\section{Introduction}
\noindent Estimating a probability density from a given set of moments known as the closure problem, naturally arises by representing a high-dimensional system with only a few moments. This inverse problem is ill-posed in general, and thus regularization/regression has to be pursued. In practice two frameworks have been developed: regression on the probability density versus regression on the logarithm of the probability density. The former includes orthogonal expansion techniques such 
as Hermite/Grad type expansions \cite{schwartz1967estimation,grad1949kinetic,struchtrup2003regularization} besides quadrature methods \cite{fox2009higher}. The latter leads to the family of Maximum Entropy Distributions (MEDs) \cite{dreyer1987maximisation,levermore1996moment}. The MED is defined by maximizing an entropy functional of the distribution, subject to the given moment constraints. Regularizing the closure problem by maximizing the Shannon entropy is motivated by both physical and information theoretic considerations. The physical motivation relies on the Boltzmann H-theorem, whereas the latter is linked to the least-bias estimators. MEDs have been employed in various settings as diverse as natural language processing \cite{och2002discriminative}, image/signal processing \cite{gull1984maximum,basseville1989distance}, geoscience \cite{ulrych1975maximum}, rarefied gas dynamics \cite{schaerer2017efficient}, solid state physics \cite{drabold1993maximum,turek1988maximum} and climate forecast \cite{schneider1999conceptual}.
However besides theoretical difficulties \cite{hauck2008convex}, the use of Maximum-Entropy distributions has been restricted due to numerical challenges.\\ 
\noindent Following standard steps of the method of Lagrange multipliers, finding the MED reduces to computing the Lagrange multipliers arising from moment constraints  \cite{kapur1989maximum}. Although the well-posedeness of such an optimization problem has been shown for bounded domains and realizable moments  
\cite{tagliani1999hausdorff,mead1984maximum,khinchin2013mathematical}, in practice expensive iterations have to be employed for finding Lagrange multipliers.
Commonly used iterative approaches are based on the gradient descent, Newton's method and the adaptive basis method. For invertable and Lipschitz continuous Hessians, Newton's method provides the fastest convergence. However since those conditions are not guaranteed in the considered setting, the adaptive basis method is suggested \cite{abramov2007improved,abramov2009multidimensional}. 
\\ 
\noindent As a numerically efficient alternative, here we reset the problem of finding the Lagrange multipliers to a Bayesian inference framework. The idea is to express the mapping from moments to Lagrange multipliers by a Gaussian Process (GP). 
Since computing moments for a given set of Lagrange multipliers is simple and cheap, the training data set can be obtained in a straight-forward way. Therefore, the hyperparameters of the considered GP prior are found by maximizing the log-likelihood over the training data set.
Once the hyperparameters are found, the Lagrange multipliers for a new set of moments can be inferred by conditioning the constructed multivariate Gaussian distribution \cite{murphy2012machine}.
\\ 
The motivation behind our approach is purely computational. Observe that all heavy computations including generating training data, finding an appropriate kernel, the Cholesky factorization of the covariance matrix and fitting the hyperparameters are done up-front (offline). For simulations, evaluation of the GP regression is done via a simple backward substitution. 
\\ 
Following the objective of constructing accurate GP estimators for the Lagrange multipliers of MED, the remainder of this manuscript is structured as the following.
First in \S~\ref{sec:reviews}, a short review of MED besides an iterative approach for computing the Lagrange multipliers are presented. Furthermore, a short description of the GP regression is provided.
Then in \S~\ref{sec:training}, training of the GP regression is pursued, where several kernels such as radial basis and Mat{\`e}rn family are evaluated for our problem setting.  Section \ref{sec:results} deals with the assessment of the devised GP-accelerated MED. As the first test case, the accuracy of the fitted GP in predicting bi-modal distributions is studied in \S~\ref{sec:testcase1_bi}. In \S~\ref{sec:noisy_bi}, robustness of the GP regression is tested by predicting MED for moments obtained from noisy bi-modal distributions. Then \S~\ref{sec:transition_BGK} {\color{hg2}and \S~\ref{sec:sol_to_boltz}} focus on relaxation of non-equilibrium distributions, governed by Bhatnagar-Gross-Krook (BGK) \cite{bhatnagar1954model}  and Boltzmann equations, respectively. At the end, a conclusion and an outlook for future studies are given in \S~\ref{sec:conclusions}.

\section{Methods}
\label{sec:reviews}
\noindent  In the following, first the MED framework is reviewed and the problem statement is refined. Next, a short description of the GP regression is presented.
\subsection{Review of Maximum Entropy Problem}
\label{sec:rev_ME}
\noindent  Consider the set of admissible probability densities defined over measurable functions as
\begin{eqnarray}
\mathcal{P}&=&\left\{f:\mathbb{R}^l\to [0,\infty)\bigg\vert \int_{\Omega}f(x)dx=1\right\},
\end{eqnarray}
where $\Omega \subseteq \mathbb{R}^l$.
Suppose we are given a finite vector of moments $ p\in \mathbb{R}^N$ of an unknown $f( v)\in\mathcal{P}$ such that
\begin{eqnarray}
p_j = \int_\Omega f( v) \phi_j( v)  d v; \ \ \ \ \ 1\leq j \leq N~,
\end{eqnarray}
where $ \phi( v):\mathbb{R}^l\to\mathbb{R}^N$ is a vector of polynomials. Here and hence forth the subscript indices denote a component of the quantity. The goal is to approximate $f$ by some $f^{(s)}\in \mathcal{P}$ such that the (mathematical) entropy 
\begin{align}
    S[f] := \int_\Omega f \ln(f) d v,
\end{align}
is minimized while the constraints
\begin{align}
    \int_\Omega  \phi( v) f^{(s)} d v =  p
\end{align}
are satisfied. Since $S[f]$ is convex and the constraints are linear, the solution of the above minimization problem is unique, once it exists. To leave out pathological cases \cite{hauck2008convex}, we focus on a bounded domain $\Omega$, for which the minimization problem is well-posed for realizable moments. Using the method of Lagrange multipliers we get  
\begin{align}
    C^{ \lambda}_N [f^{(s)}] := 
    \int_\Omega f^{(s)} \ln(f^{(s)}) d v
    - \sum_{j=1}^N \lambda_j\left( \int_\Omega f^{(s)} \phi_j d  v- p_j \right),
    \label{eq:obj_max_entropy}
\end{align}
which has its extremum at
\begin{align}
    f_N^{ \lambda} ( v) = Z_\lambda^{-1}\ {\exp{\left(-\sum_{j=1}^N\lambda_j \phi_j\right)}} ~,
\end{align}
where $Z_\lambda$ is the normalization factor \cite{kapur1989maximum}. By inserting $f_N^\lambda$ into the constraints, the Lagrange multipliers $\lambda(p)$ can be computed. However it is more convenient to consider the dual formulation which provides an unconstrained convex minimization for Lagrange multipliers as
\begin{flalign}
\label{eq:dual}
\lambda(p)&=\argmin_{\lambda^*\in \mathbb{R}^N}\left\{C(\lambda^*;p)\right\},\\
\text{where}\ \ \ \ \ C(\lambda^*;p)&:=Z_{\lambda^*}-\sum_{j}\lambda^*_jp_j~.
\end{flalign}
\noindent Hence the maximum entropy regularization, reduces the closure problem to computing $\lambda(p)$ from Eq.~\eqref{eq:dual}. 
As a direct solution of the dual problem, the standard Newton's method for finding the Lagrange multipliers are reviewed in the following.
Let $H(\lambda)$ and $g(\lambda)$ be the Hessian and the gradient of the objective function in Eq.~\eqref{eq:dual}, respectively. Following Newton's method \cite{schaerer201735}, the estimated Lagrange multipliers $\lambda^{\color{ms2}(n)}$ at step $n$, are updated by solving the linear system
\begin{align}
  \sum_{j=1}^N H_{ij}(\lambda^{\color{ms2}(n)}) \Delta {\lambda{\color{ms2}^{(n)}_j}} =  g_i(\lambda^{\color{ms2}(n)})
  \label{eq:direct_system}
\end{align}
for $\Delta \lambda^{\color{ms2}(n)}$. {\color{hg2} After random initialization of the Lagrange multipliers, they get updated according to} 
\begin{eqnarray}
\lambda{\color{ms2}^{(n+1)}_i}&=&\lambda{\color{ms2}^{(n)}_i}+\beta^{\color{ms2}(n)}\Delta \lambda{\color{ms2}^{(n)}_i}.
\label{eq:update_lambda_direct}
\end{eqnarray}
Here $\beta$ is a damping factor {\color{hg2} and is chosen such that the cost function decreases efficiently. The damping $\beta^{(n)}$  is set to the largest value of the power series $\{s^{k}\}_{k=0}^{N_s}$ with $s\in (0,1)$ that guarantees the Armijo's rule
\begin{flalign}
C(\lambda^{(n)} + \beta^{(n)} \Delta \lambda^{(n)};p)< C(\lambda^{(n)};p) + c\beta^{(n)} (\Delta \lambda^{(n)},g(\lambda^{(n)})),
\label{eq:armijo_rule}
\end{flalign}
\noindent  where $(.,.)$ indicates the dot product of vectors. Note that the values of $c$, $s$, and $N_s$ need to be tuned appropriately. For our study, we set  $c=10^{-4}$, $s=1/2$ and $N_s=30$. A pseudocode describing this direct approach with corresponding values of the free parameters are provided in algorithm~\eqref{alg:lamb_1D_direct} \cite{schaerer201735,wolfe1969convergence,wolfe1971convergence}}. Although $H$ is symmetric-positive-definite, it can become ill-conditioned which can be coped with by using an adaptive basis \cite{alldredge2014adaptive}. For example in \cite{schaerer201735}, {\color{ms1_rv1}Hermite polynomials are employed as the basis} in order to keep the Hessian matrix close to a diagonal one. A more general approach which generates a diagonal Hessian for an arbitrary probability density is followed in \cite{abramov2007improved,abramov2009multidimensional}. Yet high computational costs can become a limiting factor for this fully adaptive basis methodology.

\begin{algorithm}
\color{hg2}
\caption{\color{hg2}Direct approach to find Lagrange multipliers given the moments $p\in \mathbb{R}^N$}
\label{alg:lamb_1D_direct}
\begin{algorithmic}
\State{Set $n=0$ and sample $\lambda^{(n)}$ uniformly from $[-0.1,0.1]^N$} 
\State{Set the tolerance $\epsilon=10^{-10}$}
\While{$C(\lambda^{(n)};p)>\epsilon$ }
\State{Compute Hessian and gradient of the cost function $C(\lambda^{(n)};p)$}
\State{Solve the linear system in Eq.~\eqref{eq:direct_system} for $\Delta \lambda^{(n)}$}
\State{Find the largest $\beta^{(n)}$ that satisfies Armijo's rule \eqref{eq:armijo_rule}}
\State{Compute the new guess $\lambda^{(n+1)}$ from Eq.~\eqref{eq:update_lambda_direct}}
\State{Increment n}
\EndWhile
\\
\Return $ \lambda^{(n)}$
\end{algorithmic}
\end{algorithm}

\subsection{Gaussian Process Regression}
\label{sec:rev_GP}
\noindent  \sloppy The high computational intensity of the direct iterative approach for solving the dual problem \eqref{eq:dual}, motivates  alternative methods. Here we focus on a data-driven approach based on GP. Let us first review the main idea behind GP based regressions. Suppose $\Psi(x):\mathbb{R}^N\to\mathbb{R}^N$ is an unknown map,
yet we have access to evaluations $\{\Psi(x^{(j)})\}_{j=1}^M$ at some data points $\mathcal{D}=\{x^{(1)},x^{(2)},...,x^{(M)}\}$. Note that the superscript index denotes the corresponding data batch. Therefore the regression problem addresses estimating $\Psi(x)$ from the given $\{x^{(j)},\Psi(x^{(j)})\}_{j=1}^M$. Consider a positive semi-definite (PSD) kernel function $\mathcal{K}(x,x^\prime):\mathbb{R}^N\times\mathbb{R}^N\to[0,\infty)$, then the GP regression sets forth
\begin{eqnarray}
\label{eq:GP-general}
\tilde{\Psi} \sim \mathcal{GP}(0,\mathcal{K})
\end{eqnarray}
as an approximation of $\Psi$. Here $\mathcal{GP}$ denotes a random process whose distribution for a set of points is a joint normal with the covariance being the Gram matrix associated with $\mathcal{K}$. The merit of a regression of the type \eqref{eq:GP-general} can be addressed from different perspectives. More relevant to our setting, it can be shown that the conditional expectation $\mathbb{E}[\tilde{\Psi}|\tilde{\Psi}(x^{(j)})=\Psi(x^{(j)}), \forall x^{(j)}\in \mathcal{D}]$ provides an optimal recovery of $\Psi$ in the sense of the relative error induced by the corresponding Reproducing-Kernel-Hilbert-Space \cite{owhadi2019kernel}. In practice, we work with parametrized kernels $\mathcal{K}_{\Theta}$, where the hyperparameters embedded in $\Theta$ are found by maximizing the log-likelihood \cite{Rasmussen2006}. Furthermore, we construct the GP regressions component-wise. Hence we evaluate the hyperparameters for every $\tilde{\Psi}_i\ (i=1,...,N)$,  separately. \\ 
Several PSD kernel functions $\mathcal{K}$ have been introduced in the literature, see e.g. \cite{Rasmussen2006}. Here we consider the radial basis function (RBF) along with Mat\'{e}rn's family for each component $i,j
\in
\{1,...,N\}$ 
we have
\begin{align}
  \label{eq:RBF}  \mathcal{K}_{\Theta_{i}}^{\text{RBF}}( x, {x^\prime}) &= \sigma_i \exp\left(- r_i^2/2 \right)~,\\
    \mathcal{K}_{\Theta_{i}}^{\text{Mat\'{e}rn(12)}}( x, {x^\prime}) &= \sigma_i \exp(- r_i )~,\\
    \mathcal{K}_{\Theta_{i}}^{\text{Mat\'{e}rn(32)}}(x, {x^\prime}) &= \sigma_i (1+\sqrt{3}r_i)\exp(- \sqrt{3}r_i ) \quad \text{and} \\
    \mathcal{K}_{\Theta_{i}}^{\text{Mat\'{e}rn(52)}}(x, {x^\prime}) &= \sigma_i (1+\sqrt{5}r_i + \frac{5}{3}\sqrt{r_i}{\color{ms1})}\exp(- \sqrt{5}r_i )~.
\end{align}
Note that $r^2_i=\sum_{j}L^{-1}_{ij}(x_j-x_j^\prime)^2$, where the positive-definite-matrix $L_{N \times N}$ 
encodes a characteristic length-scale.
For each component $i\in\{1,...,N\}$, the hyperparameters $\Theta_{i}=\{\sigma_i,L^{-1}_{i1},...,L^{-1}_{iN}\}$ can be found by maximizing the log-likelihood 
\begin{align}
    \ln & \left(   \tilde{f}\left (\tilde{\Psi}_i(x)\,|\,   x\in \mathcal{D} \right)\right) = -\frac{1}{2} \ln\left({\color{ms1}\det(\mathcal{K}_{\Theta_{i}}(x, {x^\prime}))}\right) \nonumber \\
    &{\color{ms1}-\frac{1}{2}} {{\Psi}_i}^T(x) \mathcal{K}_{\Theta_{i}}(x, {x^\prime})^{-1} {\Psi}_i(x^{\color{ms1}\prime}) -\frac{M}{2} \ln(2\pi) ,
    \label{eq:likelihood}
\end{align}
where $\tilde{f}$ denotes the probability density of $\tilde{\Psi}_i$ conditioned on the training points.
The Broyden-Fletcher-Goldfarb-Shanno algorithm (BFGS) is used in this study to find the local minimum with respect to the hyperparameters \cite{nocedal2006numerical}. It can be shown that the global minimum is attained as more data points are deployed \cite{gyorfi2006distribution}.
Once the kernel function $\mathcal{K}$ and its hyperparameters are set, one can evaluate the distribution of $\tilde{\Psi}$ at an arbitrary input point {\color{hg1} $x^*\in \mathbb{R}^N$}. Let $x$ be composed of the training points, therefore
\begin{align}
\left(\tilde{\Psi}_i ({x^*}) \bigg \vert  \tilde{\Psi}_i(x)=\Psi_i(x)\right)
\sim
\mathcal{N}(\bar{m}_i,\bar{\Sigma}_i),
\end{align}
where {\color{hg1} $\mathcal{N}(\bar{m}_i,\bar{\Sigma}_i)$ is a 
normal distribution with mean}
\begin{align}
\bar{m}_i &= \mathcal{K}_{\Theta_i}( {x^*}, {x^{\prime}})   \mathcal{K}_{\Theta_i}( x, {x^{\prime}})^{-1} \Psi_i(x)
\label{eq:pred_mean}
\end{align}
{\color{ms1}
and 
variance}
\begin{align}
\bar{\Sigma}_i&=\mathcal{K}_{\Theta_{i}}(x^*, x^*) -   \mathcal{K}_{\Theta_i}( x^*, x) \mathcal{K}_{\Theta_i}( x, x^{\prime})^{-1}  \mathcal{K}_{\Theta_i}( x^*, x)~{\color{ms1}.}
\label{eq:pred_var}
\end{align}
{\color{hg1} Note that $\tilde{m}$ and $\tilde{\Sigma}$ indicate posterior estimates of the mean and the variance, respectively.}
 Since the inversions appearing in Eqs.~\eqref{eq:pred_mean} and \eqref{eq:pred_var} only include the training points, the corresponding computations can be done up-front. {\color{hg1}Therefore computational advantage is gained, as only matrix-vector multiplication is needed for predictions.} Although more efficient GP models such as sparse GP \cite{snelson2006sparse} could be pursued,  in this study we adopt the straight-forward GP regression model available on GPflow \cite{GPflow2017}.
\section{Training Gaussian Process}
\label{sec:training}
\floatpagestyle{plain}
 \noindent  In this section, constructing GP maps for Lagrange multipliers are pursued. The performance of several covariance functions besides accuracy of the GP regression close to realizability limit are assessed.
\label{sec:res_ME1D}
\subsection{Initializing data set}
\label{sec:data_set}
\noindent  To construct a regression on the Lagrange multipliers as a map from moments, we need to construct a data set. Since the inverse map is cheaper to evaluate, the main idea here is to compute the moments based on a set of Lagrange multipliers. In order to do that we need to introduce a domain for Lagrange multipliers of the form $\Lambda=[\lambda_{\min},\lambda_{\max}]^N$ to sample from. Furthermore, we need to have a boundary for values of the moments i.e. $\Omega_p=\prod_{i} [p_{i,\min},p_{i,\max}]$. Since all scenarios can be shifted and scaled to a reference with zero mean and unity variance, finally we only include the data points corresponding to zero mean and unity variance MEDs.
{\color{hg2}{
First, $\{\tilde{\lambda}_i\}_{i=1}^N$ are uniformly sampled from $\Lambda$ resulting in a trial density ${f}_N^{\tilde{\lambda}}$}}. The mean $\mu$ and the variance $\sigma^2$ are computed from $f_N^{\tilde{\lambda}}$ using Gaussian-quadrature. In order to find the corresponding Lagrange multipliers that guarantee zero mean and unity variance, we make use of the coordinate transformation $v^\prime=(v-\mu)/\sigma$. {\color{hg2}{Let $f_{N}^\lambda$ be the density with zero mean and unity variance.}} Observe that by equality of measures {\color{hg2}{and assuming $v\in \mathbb{R}$}} we get 
\begin{eqnarray}
\label{eq:trans}
f_{N}^{\lambda}(v^\prime)&=&\sigma f_{N}^{\tilde{\lambda}}\left(\sigma v^\prime+\mu\right).
\end{eqnarray}
Using the binomial expansion, it is straight-forward to find that
\begin{eqnarray}
\label{eq:update-lambda}
\lambda_i=\sigma^i\tilde{\lambda}_i+\sum_{j=i+1}^N \tilde{\lambda}_j {j \choose i} \sigma^i \mu^{j-i}~; \ \ \ i\in\{1,...,N\}
\end{eqnarray}
ensures Eq.~\eqref{eq:trans}. {\color{hg2}{Yet since we deal with bounded $v\in \Omega$, an iterative scheme with the initial guess given by 
Eq.~\eqref{eq:update-lambda} is employed to ensure zero mean and unity variance. Algorithm~\eqref{alg:lamb_1D} is introduced in order to create the data set.}}

\begin{algorithm}
{\color{hg2}
\caption{\color{hg2}Generating $( \lambda, p) $ with $\int_{\Omega}v f_N^\lambda dv=0$ and $\int_{\Omega}v^2 f_N^\lambda dv=1$ given the moment space $\Omega_p$, and sample space $\Lambda$ for Lagrange multipliers}  
\label{alg:lamb_1D}
\begin{algorithmic}
\State {Set  $p^\prime=(0,...,0)^T\in\mathbb{R}^{N}$ with $N=\dim(\Omega_p)$} 
\State{Set the tolerance $\epsilon=10^{-10}$}
\While{$p^\prime\notin \Omega_p$}
\State{Sample ${\tilde{\lambda}}$ uniformly from
$\Lambda$
}
\State{Compute $\mu=\int_{\Omega}v f_N^{\tilde{\lambda}} dv$ and $\sigma^2=\int_{\Omega}v^2 f_N^{\tilde{\lambda}} dv$}
\While{$\mu>\epsilon$ or $|\sigma-1|>\epsilon$}
\State{Compute $ \lambda$ according to Eq.~\eqref{eq:update-lambda}}
\State{Update $\mu=\int_{\Omega}v f_N^{{\lambda}} dv$ and $\sigma^2=\int_{\Omega}v^2 f_N^{{\lambda}} dv$}
\State{$\tilde{\lambda} \gets \lambda$} 
\EndWhile
\State{Update $p^\prime_i=\int_{\Omega} v^if_N^{{\lambda}}dv$ for $1\leq i \leq N$ }
\EndWhile
\State{$p \gets p^\prime$} \\
\Return $\lambda$ and $p$
\end{algorithmic}
}
\end{algorithm}

\noindent  For the training, we consider $N\in\{4,6,8\}$
and
{\color{ms2}n}umerical integrations are carried out using Gaussian-quadrature with roughly $20$ points.
{\color{hg2} The sample space for the Lagrange multipliers has been chosen carefully after a trial and error on the outcome moments obtained from the algorithm \eqref{alg:lamb_1D}. As shown in Fig.~\ref{fig:expansion}, the generated data points using the uniform sample space of $\Lambda=[-b,b]^{N}$ with $b\in\{1,10 \}$,  suggests that the tail of the moment distribution becomes longer as $b$ increases. This implies that MED with a larger $b$ is better equipped to capture rare events. 
The training points are generated by setting $b=10$ and only keeping data points whose moments lie in the space 
$\Omega_p = [-\epsilon,\epsilon]\times [1-\epsilon,1+\epsilon] \times [-1,1]\times[1,{\color{ms1_rv1}4}]\times[-4,4]\times[1,15]\times[-25,1]\times[1,110]$.
\begin{figure}
  \centering
  \begin{subfigure}[b]{0.48\columnwidth}
  \includegraphics{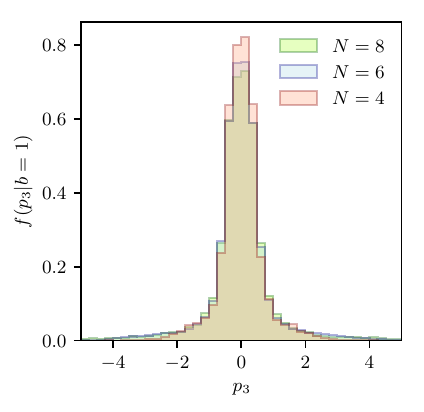}
  \caption{}
  \end{subfigure}
  \begin{subfigure}[b]{0.48\columnwidth}
  \includegraphics{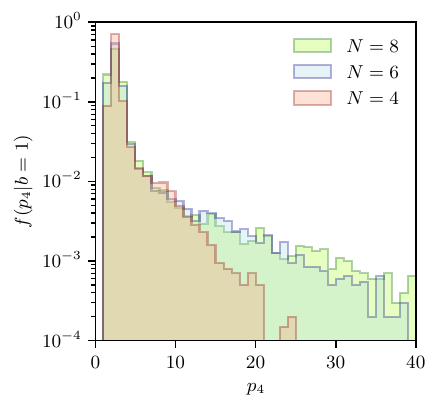}
  \caption{}
  \end{subfigure}
  \begin{subfigure}[b]{0.48\columnwidth}
  \includegraphics{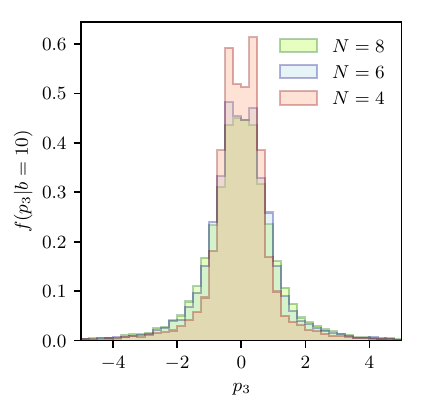}
  \caption{}
  \end{subfigure}
  \begin{subfigure}[b]{0.48\columnwidth}
  \includegraphics{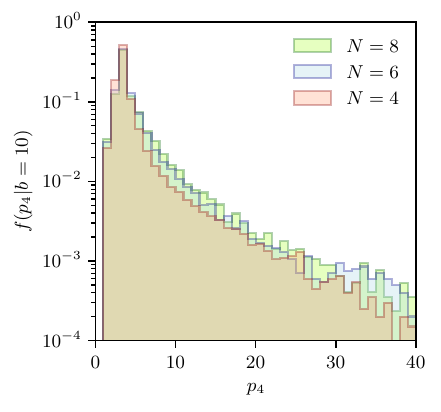}
  \caption{}
  \end{subfigure}
  \caption{\color{hg2}Probability density function of the moments $p_3$ and $p_4$ obtained from $10^4$ data points using algorithm \eqref{alg:lamb_1D}. The sample space $\Lambda=[-b,b]^{N}$  for the Lagrange multipliers varies with   $N\in\{4,6,8 \}$  and $b\in\{1,10 \}$.}
  \label{fig:expansion}
\end{figure}
}
\subsection{Pre-treatment of data set}
\noindent Every $(\lambda^{(k)}_i,p^{(k)}_j)$ component of the data set can have significant variations passing through different batches of $k\in\{1,...,M\}$ (with $M=1000$ for our data set). 
We follow the common recipe in data-driven methodologies which includes scaling and shifting of every data point $(\lambda^{(k)}_i,p^{(k)}_j)$ by the standard-deviation and the average computed over $N$ batches of the particular $(i,j)$ component, respectively. Note that this does not have to be carried out for $p_1$ and $p_2$, since they have fixed values already. 
\subsection{Kernel comparison}
\begin{figure}
  \centering
  \begin{subfigure}[b]{0.48\columnwidth}
  {\includegraphics[ angle =0]{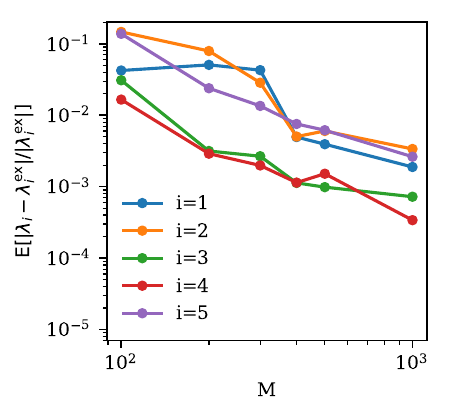}}
    \caption{}
  \end{subfigure}
  \  
  \begin{subfigure}[b]{0.48\columnwidth}
  {\includegraphics[ angle =0]{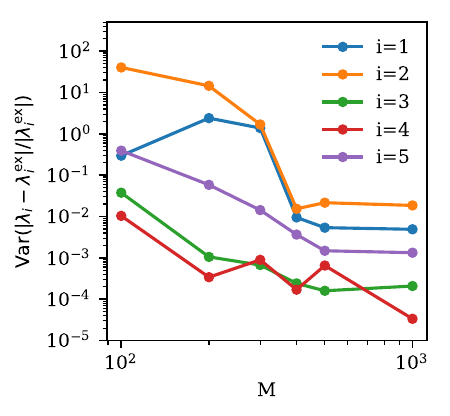}}
    \caption{}
  \end{subfigure}
    \caption{Expectation and variance of the relative error in predicting $\lambda$s using RBF. {\color{hg1}Here, $\lambda^{\mathrm{ex}}\in \mathbb{R}^N$ with $N=6$ indicates the exact solution of MED taken from untrained subset of the data set. The statistics are performed over $2000$ testing points.}}
  \label{fig:convergence1}
\end{figure}

\noindent {\color{hg1} We consider MED with $N=6$ moments as the target distribution, where the appropriate kernel and number of training data points $M$ should be found.} First, let us consider the radial basis function (RBF) for the kernel choice. Once the hyperparameters of Eq.~\eqref{eq:RBF} are found via maximizing the log-likelihood given by Eq.~\eqref{eq:likelihood}, the accuracy of predictions over unseen data is investigated. As shown in Fig.~\ref{fig:convergence1}, by increasing the number of data points $M$ in the training set, the expectation and the variance of the relative error decay using the GP regression.

\begin{figure}[htbp]
  \centering
  \begin{subfigure}[b]{0.48\columnwidth}
  \includegraphics{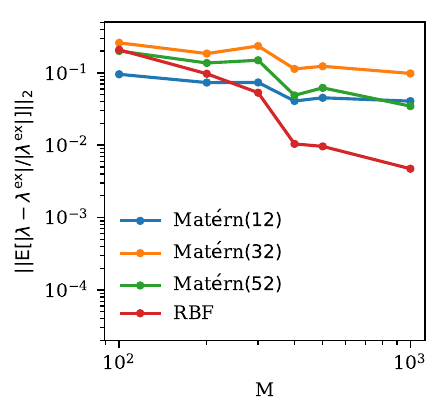}
  \caption{}
  \end{subfigure}
 \begin{subfigure}[b]{0.48\columnwidth}
   
  \includegraphics{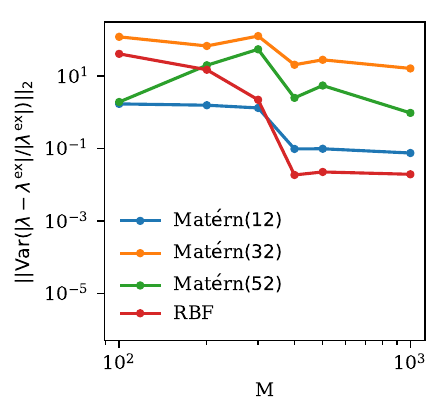}
  \caption{}
  \end{subfigure}
  \caption{Convergence comparison using different kernels. The $L^2$-norm of the expectation and the variance of the relative error are shown. {\color{hg1}Here, $\lambda^{\mathrm{ex}}\in \mathbb{R}^N$ with $N=6$ indicates the exact solution of MED taken from untrained subset of the data set. The statistics are performed over $2000$ testing points.}}
  \label{fig:convergence2}
\end{figure}

\noindent For comparison, several kernels from the Mat\'{e}rn family of functions, i.e.  Mat\'{e}rn(12), Mat\'{e}rn(32) and Mat\'{e}rn(52), have been tested here for the training step. Based on our computational experiments as shown in  Fig.~\ref{fig:convergence2}, RBF provides a better estimation for this data set. 

\subsection{Cost of data generation and training}
\begin{figure}
  \centering
  \begin{subfigure}[b]{0.48\columnwidth}
  \includegraphics{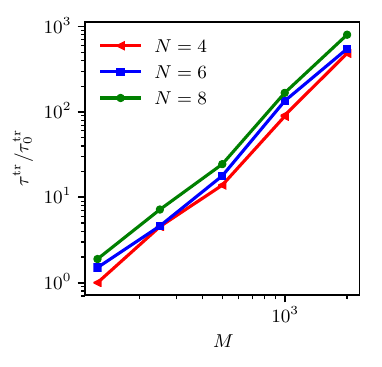}
  \caption{}
  \end{subfigure}
  \begin{subfigure}[b]{0.48\columnwidth}
  \includegraphics{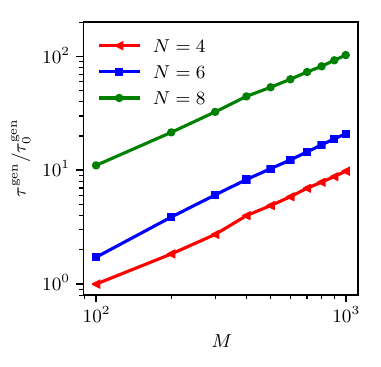}
  \caption{}
  \end{subfigure}
  \caption{\color{hg2} Averaged execution time $\tau^{\mathrm{gen}}$ for tuning hyperparameters and $\tau^{\mathrm{tr}}$ for generating the data set, are depicted at left and right, respectively. Here $N$ denotes the number of moments and $M$ the number of employed data points. Execution times are normalized by the computational time corresponding to the case of  $N=4$ and $M=100$.}
  \label{fig:tr_gen_cost}
\end{figure}

\noindent Although the described  algorithm\eqref{alg:lamb_1D} is straight-forward, the iterations on $\tilde{\lambda}$ required to ensure $p\in\Omega_p$ besides zero mean and unity variance, can become costly. As it can be seen in Fig.~\ref{fig:tr_gen_cost}, the computational time for generating the data set $\tau^{\mathrm{gen}}$ scales almost linearly with the number of data points. However, the cost of generating relevant data points increases as more moments are considered. \\
\noindent The GP hyperparameters are trained by maximizing the log-likelihood, as explained in \S~\ref{sec:rev_GP}. The execution time for training $\tau^{\mathrm{tr}}$ shown in Fig.~\ref{fig:tr_gen_cost} includes cost of the Cholesky factorization besides optimizing the hyperparameters with a tolerance $10^{-6}$.

\subsection{Accuracy in the limit of realizability}
\begin{figure}
  \centering
  \includegraphics{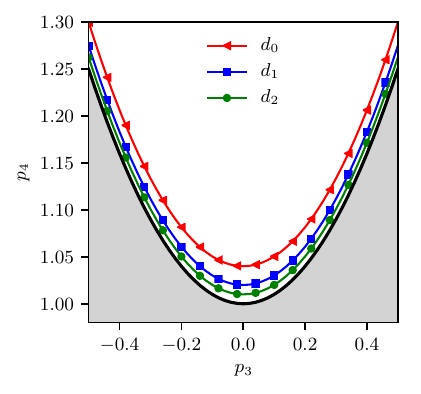}
  \includegraphics{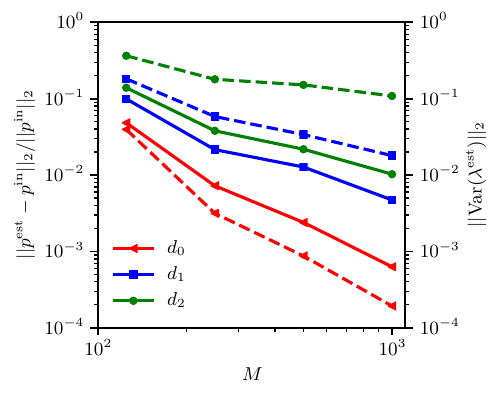}
  \caption{\color{hg1} Left: Input points depicted in $(p_3,p_4)$ plane with solid black curve on $p_4=p_3^2+1$, Right: Average of moments relative error using data-driven MED for input moments $p^{\mathrm{in}}\in D_p^{\mathrm{test}}$, and variance of the predicted Lagrange multipliers shown in solid and dashed lines, respectively}
  \label{fig:realizability}
\end{figure}

\noindent Here, we investigate the accuracy of the trained GP at the limit of moment realizability. Let us consider MED with $N=4$ moments along with RBF as the kernel function. Following \cite{mcdonald2013affordable,akhiezer1965classical}, the moment problem is physically realizable for $N=4$, if the sufficient condition
\begin{flalign}
p_4 \geq p_3^2+1~
\end{flalign}
holds for the standardized moments $p\in\mathbb{R}^4$.
In order to show accuracy of the trained GP at the points near the limit $p_4=p_3^2+1$, we investigate the GP predictions for a set of standardized input moments
\begin{flalign}
D_{p}^{\mathrm{test}}=\Big \{
\begin{psmallmatrix}
0\\ 
1\\ 
\alpha\\ 
\alpha^2+1+d
\end{psmallmatrix}
\Big| \alpha=\alpha_{\text{min}}+ih,\  i=1,...,N_{\mathrm{test}},\  h=({\alpha_{\text{max}}-\alpha_{\text{min}}})/{N_{\mathrm{test}}}
\Big \}
\end{flalign}
where $d\in\{0.04,0.02,0.01 \}$, $\alpha_{\mathrm{max}}=-\alpha_{\mathrm{min}}=0.5$,  and $N_{\mathrm{test}}=100$.
As shown in Fig.~\ref{fig:realizability}, by decreasing $d$, the relative error and the variance of predictions increase. On the other hand, as expected, the accuracy improves by increasing the number of training points $M$. Therefore this investigation suggests that the GP regression for MED becomes less reliable once moments close to the realizability border are encountered.

\begin{figure}[htbp]
  \centering
  \includegraphics{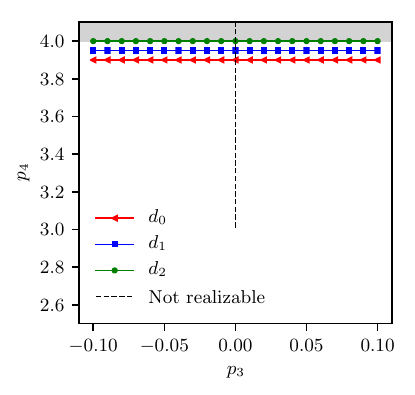}
  \includegraphics{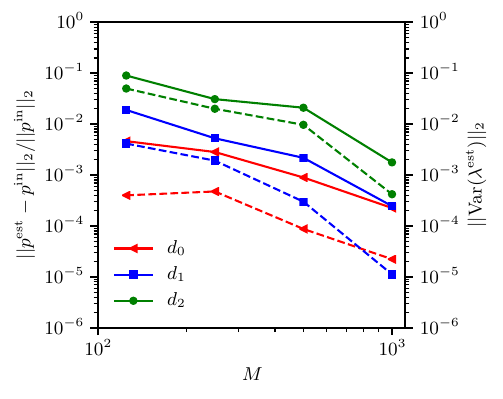}
  \caption{\color{ms1_rv1} Left: Input points depicted in $(p_3,p_4)$ plane with black dashed line depicting $p_3=0$ with $p_4>3$, Right: Average of moments relative error using data-driven MED for input moments $p^{\mathrm{in}}\in U_p^{\mathrm{test}}$, and variance of the predicted Lagrange multipliers shown in solid and dashed lines, respectively}
  \label{fig:realizability_up2}
\end{figure}

\begin{figure}[htbp]
  \centering
  \includegraphics{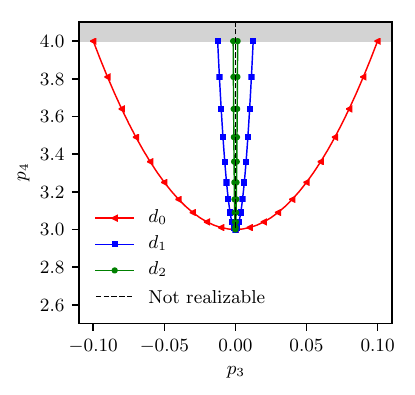}
  \includegraphics{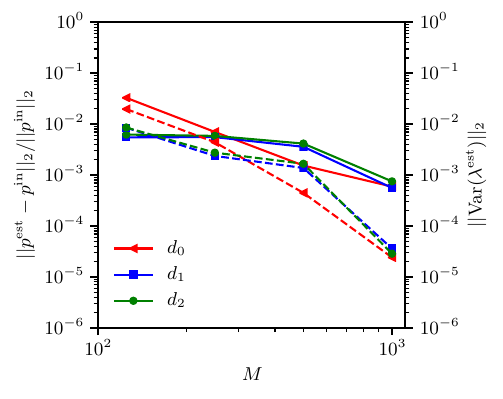}
  \caption{Left: Input points depicted in $(p_3,p_4)$ plane with black dashed line depicting $p_3=0$ with $p_4>3$, Right: Average of moments relative error using data-driven MED for input moments $p^{\mathrm{in}}\in S_p^{\mathrm{test}}$, and variance of the predicted Lagrange multipliers shown in solid and dashed lines, respectively}
  \label{fig:realizability_up}
\end{figure}

\noindent Moreover, the upper limit of realizable moments is investigated here. First, since data points with 4th order moment  $p_4\in[1,4]$ are considered here, let us evaluate the 
accuracy of GP in predicting MED as points of interest approaches the upper limit. Let us define  a set of points as
\begin{flalign}
U_{p}^{\mathrm{test}}=\Big \{
\begin{psmallmatrix}
0\\ 
1\\ 
\beta\\ 
4-d
\end{psmallmatrix}
\Big| \beta=\beta_{\text{min}}+ih,\  i=1,...,N_{\mathrm{test}},\  h=({\beta_{\text{max}}-\beta_{\text{min}}})/{N_{\mathrm{test}}}
\Big \}
\end{flalign}
where $d\in \{ 0.1, 0.05, 0\}$ and $\beta_{\mathrm{max}}=-\beta_{\mathrm{min}}=0.1$, and $N_{\mathrm{test}} = 200$ illustrate the number of testing points. As shown in the Fig.~\ref{fig:realizability_up2}, similar to the lower limit of physical realizability,  the accuracy in prediction decreases as the upper limit of moments is approached.
\\ Finally, the trained GP is tested in estimating the MED near the line $p_3=0$ with $p_4>3$ which are not realizable with MED. In order to evaluate the accuracy of MED with $N=4$ around this limit of realizability, let us take moments from the set
\begin{flalign}
S_{p}^{\mathrm{test}}=\Big \{
\begin{psmallmatrix}
0\\ 
1\\ 
\beta/d\\ 
(10\beta d)^2+3
\end{psmallmatrix}
\Big| \beta=\beta_{\text{min}}+ih,\  i=1,...,N_{\mathrm{test}},\  h=({\beta_{\text{max}}-\beta_{\text{min}}})/{N_{\mathrm{test}}}
\Big \}
\end{flalign}
where the parameters are $d\in\{1,8,64 \}$  and $N_{\mathrm{test}}=100$ indicates the number of testing points. Similar to the lower bound of realizability, it can be observed from Fig.~\ref{fig:realizability_up}  that the relative error, as well as the variance of the predictions, decrease by deploying more training points. However, the error in predictions as the point of interest approaches the upper limit of realizability, i.e., by increasing $d$, is negligible, which can be explained by having an excess of testing points near the equilibrium point, i.e., $(p_3,p_4)=(0,3)$.

\section{Results}
\label{sec:results}
\noindent  In this section, the trained GP is employed for predicting different scenarios relevant in kinetic problems. To further refine our setting, without loss of generality we restrict ourselves to a one-dimensional domain $\Omega=[v_{\text{min}},v_{\text{max}}]$. Moreover the moments are computed for the polynomials $\phi_i=v^i$, for $i\in\{1,...,N\}$. We shift and scale the coordinate such that zero mean and unity variance are obtained. After normalization, the velocity sample space is set by adopting $v_{\text{max}}=-v_{\text{min}}=10$. 
\subsection{Test case \#1: recovering bi-modal density}
\label{sec:testcase1_bi}
\begin{figure}[htbp]
  \centering
  \begin{subfigure}[b]{0.48\columnwidth}
 \includegraphics{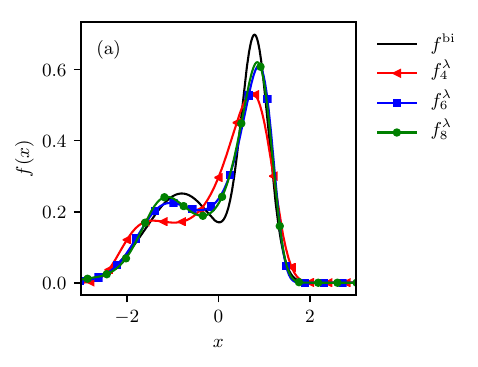}
  \caption{}
  \end{subfigure}
  \begin{subfigure}[b]{0.48\columnwidth}
  \includegraphics{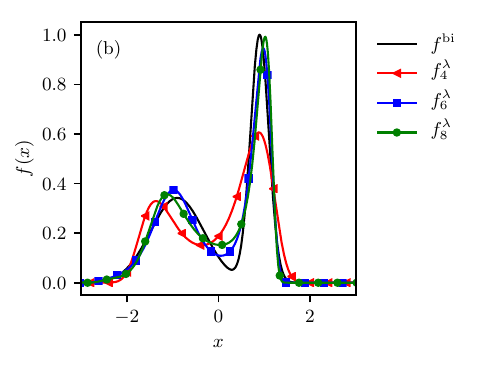}
  \caption{}
  \end{subfigure}
 \begin{subfigure}[b]{0.48\columnwidth}
   \includegraphics{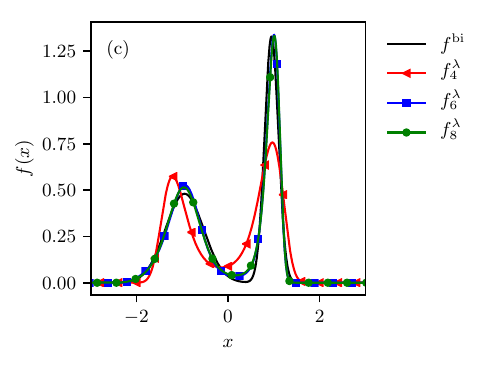}
   \caption{}
   \end{subfigure}
 \begin{subfigure}[b]{0.48\columnwidth}
  \includegraphics{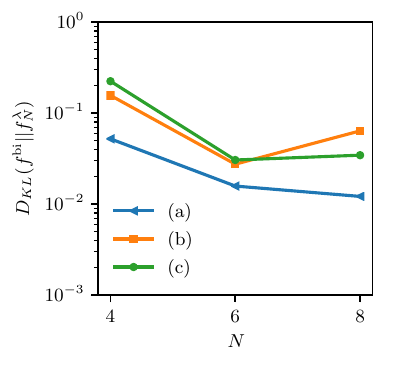}
  \caption{}
  \end{subfigure}
  \begin{subfigure}[b]{0.48\columnwidth}
    \includegraphics{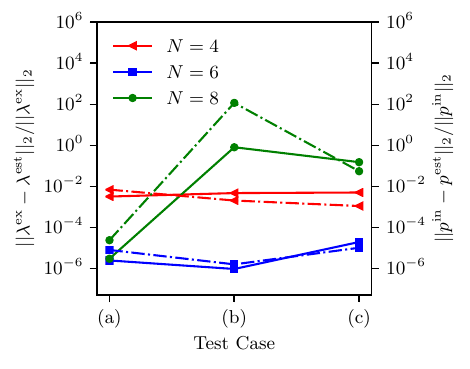}
    \caption{}
  \end{subfigure}
  \begin{subfigure}[b]{0.48\columnwidth}
    \includegraphics{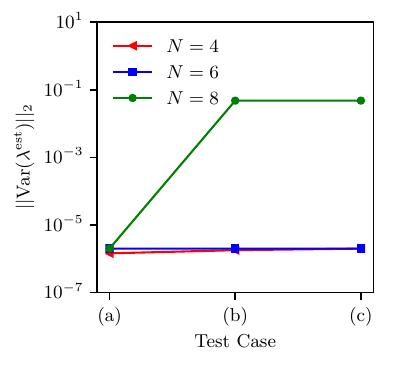}
    \caption{}
  \end{subfigure}
  \caption{Recovering bi-modal probability densities using MEDs accelerated by the GP regression with $N=4,6$ and $8$ moments. {\color{hg2} The estimated densities are shown in sub-figures (a)-(c) for test cases (a)-(c), respectively. The KL-divergence between estimated MEDs and the bi-modal distribution, relative error of the GP estimator with respect to exact values of Lagrange multipliers and  outcome moments, and variance of predictions are presented in (d)-(f), respectively.}}
  \label{fig:bimodal_distribution_var}
 \thisfloatpagestyle{empty}
\end{figure}

\noindent  Bi-modal distributions are prototype of non-equilibrium phenomena in kinetic problems. For example they show up as simplified solutions of shock waves in rarefied gas kinetics \cite{mott1951solution}. We employ the trained GP with the RBF kernel to predict the bi-modal density of the form
\begin{align}
    f^{\text{{\color{ms1}bi}}} (v|\mu_1, \sigma_1, \mu_2, \sigma_2) = \frac{1}{2} \left[f^{\mathcal{N}}(v|\mu_1,\sigma_1)+f^{\mathcal{N}}(v|\mu_2,\sigma_2) \right],
\end{align}
where $\mu_2 = - \mu_1$ and $\sigma_2=\sqrt{2 - (\sigma_1 ^ 2 + 2  \mu_1 ^ 2)}$.
Note that $f^{\mathcal{N}}(v|\mu,\sigma)$ is the normal density with the mean $\mu$ and the variance $\sigma^2$. To quantify the deviation of the estimated density from the exact one, the Kullback--Leibler divergence 
\begin{align}
    D_{KL}(f^{\text{{\color{ms1}bi}}} || f_N^\lambda) = \int_\Omega f^{\text{{\color{ms1}bi}}}(v) \ln\left(f^{\text{{\color{ms1}bi}}}(v) / f_N^\lambda(v) \right) dv
\end{align}
is used here. Three different scenarios {\color{hg1}{\{(a),(b),(c)\} corresponding to $(\mu_1,\sigma_1) \in \{ (0.8,0.3), (0.9,0.2), (0.95,0.15)\}$ are considered}}, where predictions are provided based on the GP regression with $N=4,6$ and $8$ moments. 
The results depicted in Fig.~\ref{fig:bimodal_distribution_var} show that even with $f^\lambda_4$ a good recovery is achieved. 
{\color{hg1}
Although predictions of  MED suggest that by increasing the number of moments better estimation of the bi-modal distribution can be obtained, such improvement were not observed for the predictions in the test case (b) and (c) from $N=6$ to $N=8$. This discrepancy can be explained by noticing high values for the variance of posterior for mentioned points, i.e. lack of training data near prediction points.}
As expected by merging the two modes, better agreement is obtained between the GP-accelerated MED and the bi-model one. 
 {\color{hg2}
 \\ \ \\
 To further evaluate accuracy and performance of the data-driven MED, the bi-modal test case was also studied using the standard algorithm~\eqref{alg:lamb_1D_direct}. While reasonable accuracy in predicting the exact Lagrange multipliers and their outcome moments are obtained via GP estimates  as shown in Fig.~\ref{fig:bimodal_distribution_err}, a speedup of at least two orders of magnitude compared to the direct approach is observed. The predictions are improved overall as the number of moments is increased. 
 }

{
\color{ms2}
\begin{figure}[htbp]
  \centering
     \includegraphics{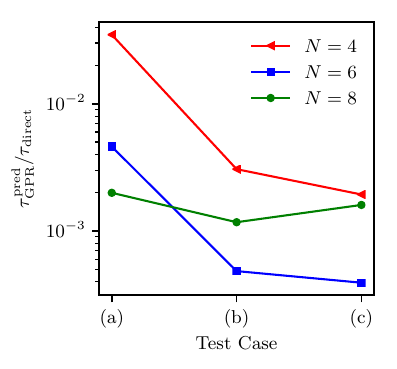}
  \caption{\color{ms2}The ratio between computational times of the GP regression and the direct approach are shown for computing the Lagrange multipliers.}
  \label{fig:bimodal_distribution_err}
\end{figure}
}

\subsection{Test case \#2: noisy bi-modal distribution}
\label{sec:noisy_bi}
\begin{figure}[htbp]
\centering
\begin{subfigure}[b]{0.48\columnwidth}
\includegraphics{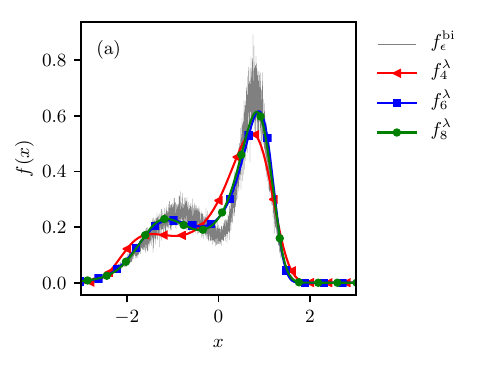}
\caption{}
\end{subfigure}
\begin{subfigure}[b]{0.48\columnwidth}
  \includegraphics{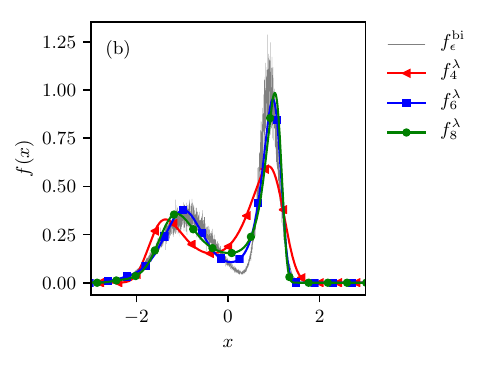}
  \caption{}
  \end{subfigure}
  \begin{subfigure}[b]{0.48\columnwidth}
  \includegraphics{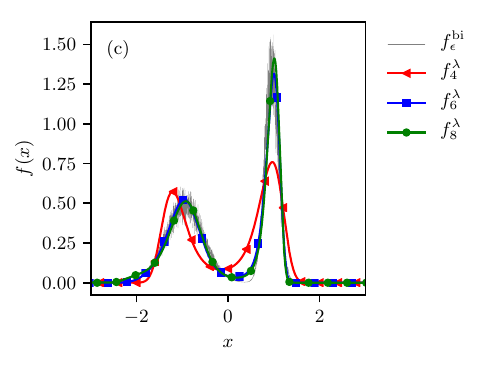}
  \caption{}
  \end{subfigure}
\begin{subfigure}[b]{0.48\columnwidth}
  \includegraphics{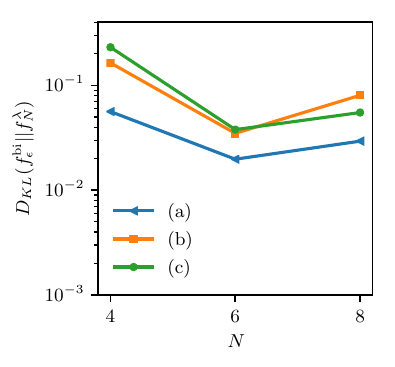}
  \caption{}
  \end{subfigure}
  \begin{subfigure}[b]{0.48\columnwidth}
  \includegraphics{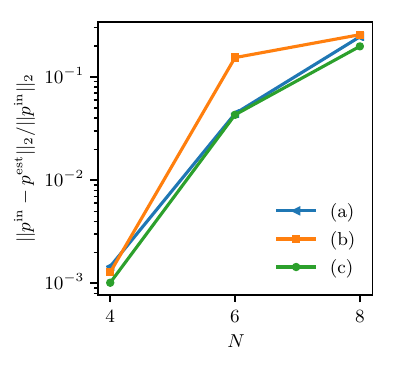}
  \caption{}
 \end{subfigure}
 \begin{subfigure}[b]{0.48\columnwidth}
  \includegraphics{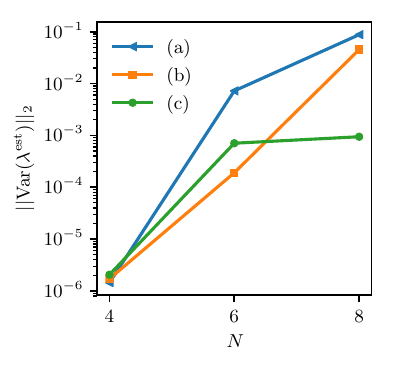}
  \caption{}
   \end{subfigure}
  \caption{\color{hg2}Estimating the noisy bi-modal distribution $f_\epsilon^{\mathrm{bi}}$ with GP-accelerated MED $f_N^\lambda$, where $N\in\{4,6,8\}$ for test cases $\{ (a), (b), (c) \}$. Probability density functions are shown in (a)-(c). Also the KL-divergence between distributions, the relative error in outcome moments and the variance of the Lagrange multipliers are shown in (d)-(f), respectively.}
  \label{fig:bi_noise}
  \thisfloatpagestyle{empty}
\end{figure}

\noindent To test the robustness of our data-driven MED estimator, here we consider a perturbed bi-modal distribution
\begin{align}
    f^{\text{{\color{hg2}bi}}}_\epsilon (v|\mu_1, \sigma_1, \mu_2, \sigma_2) = f^{\text{{\color{hg2}bi}}}(v|\mu_1, \sigma_1, \mu_2, \sigma_2)(1+\epsilon),
\end{align}
where $\epsilon$ is a random variable with the normal density $f^{\mathcal{N}}(0,0.1)$. The values of $(\mu_1,\sigma_1)$ are taken from \S~\ref{sec:testcase1_bi}. 
\\ \ \\
As depicted in Fig.~\ref{fig:bi_noise}, devised GP estimators provide reasonable performance for perturbed scenarios. Here, it can be observed that although MED with higher moments has the potential of describing more complicated distributions, the sensitivity of higher-order Lagrange multipliers to the input moments reduces the robustness.

\subsection{Test case \#3: recovering BGK relaxation}
\label{sec:transition_BGK}
\begin{figure}
  \centering
  \begin{subfigure}[b]{0.48\columnwidth}
   \scalebox{0.93}{
  \includegraphics{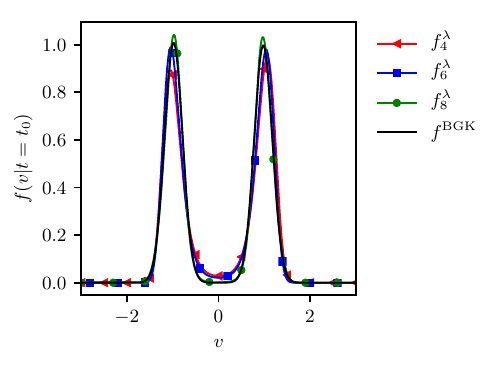}}
  \caption{}
  \end{subfigure}
 \begin{subfigure}[b]{0.48\columnwidth}
   \scalebox{0.93}{
  \includegraphics{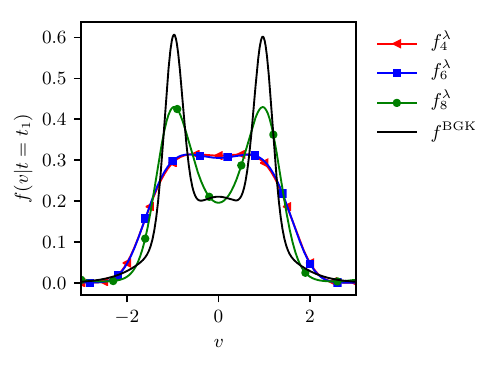}
  }
  \caption{}
  \end{subfigure}
 \begin{subfigure}[b]{0.48\columnwidth}
 \scalebox{0.93}{
   \includegraphics{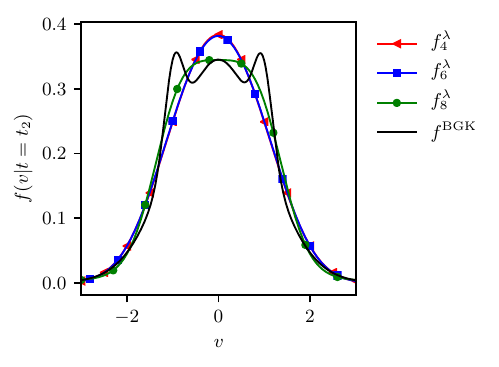}
   }
   \caption{}
   \end{subfigure}
   \begin{subfigure}[b]{0.48\columnwidth}
   \scalebox{0.93}{
 \includegraphics{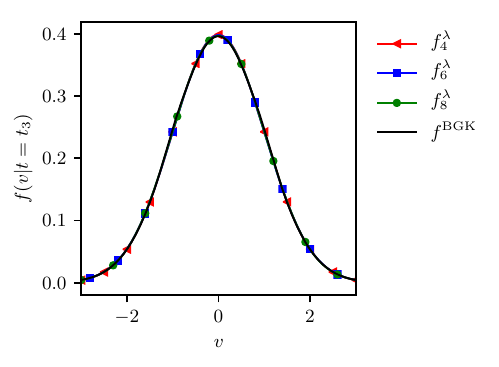}
 }
 \caption{}
   \end{subfigure}
    \caption{Capturing BGK relaxation using GP-accelerated MED  for $N=4,6$ and $8$ moments at time 
    $t\in \{  0,3,8,20 \}$.
    }
    \label{fig:transition2}
  
\end{figure}

\begin{figure}
  \centering
  \scalebox{1.0}{
  \includegraphics{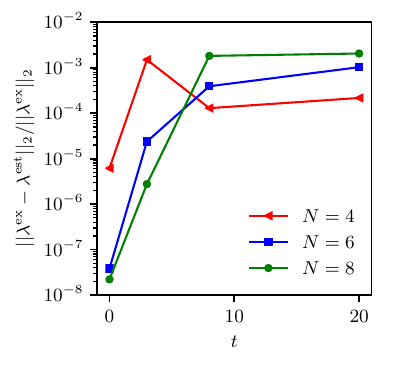}}
  \scalebox{0.95}{
  \includegraphics{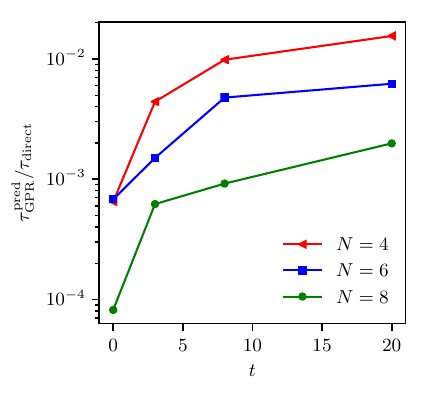}}
 \caption{Left: Relative error of the GP regression in predicting Lagrange multipliers of MED for $N=4,6$ and $8$, Right: The ratio between computational times required by direct and GP approaches for $N=4,6$ and $8$ at time $t\in \{  0,3,8,20 \}$.}
  \label{fig:transition3}
\end{figure}

\noindent This test case investigates the accuracy of the trained GP with the RBF kernel in predicting the evolution of a density  $f( v|t)$ governed by
\begin{flalign}
\frac{\partial f(v|t)}{\partial t} = \nu (f^{\mathcal{N}}(v|0,1)-f(v|t))~.
\label{eq:bgk}
\end{flalign}
The collision frequency $\nu$ controls how quick the solution reaches the equilibrium. Given an initial condition $f(v|t_0)$, the exact solution reads
\begin{flalign}
f^{\text{ex}}(v|t) = \left[1-\exp(-\nu t)\right]f^{\mathcal{N}}(v|0,1) +  \exp(-\nu t)f(v|t_0)~.
\label{eq:bgk_ex_sol}
\end{flalign}
Here, we use bi-modal normal distribution described in \S~\ref{sec:testcase1_bi} with $(\mu_1,\sigma_1) = (0.98,0.2)$ as the initial density. \\ \ \\
In order to solve Eq.~\eqref{eq:bgk} using MED, the Lagrange multipliers corresponding to the set of moments at time $t$ need to be evaluated. Applying the devised GP regression, trained for $\lambda \in \mathbb{R}^N$ with $N=4,6$ and $8$, the Lagrange multipliers are estimated.
Observe that the moments $p(t)$ can be computed analytically from Eq.~\eqref{eq:bgk}.
Therefore, at each time instant, the moments and subsequently the trained GP map, are found. 
The estimated $f^\lambda_N$ together with its moments are compared with respect to the corresponding exact solution as shown in Fig.~\ref{fig:transition2}. Here $\nu=0.25$ and time intervals are $(t_0=0,t_1=3,t_2=8,t_3=20)$ are chosen. 
\noindent 
Improvements of the estimator are clearly visible by increasing the number of moments as shown in Fig.~\ref{fig:transition3}. It is encouraging to see that even with as few moments as $N=4$, one can recover the bi-model density using the GP-estimated MED.

\subsection{Test case \#4: recovering Boltzmann relaxation}
\label{sec:sol_to_boltz}
\begin{figure}
\centering
\begin{subfigure}[b]{0.48\columnwidth}
  \includegraphics{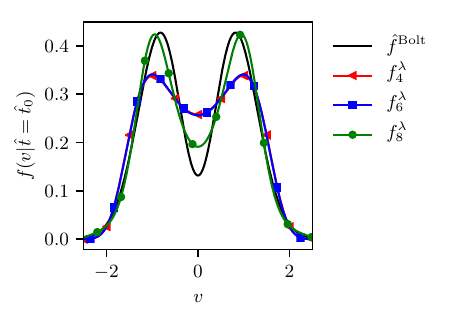}
  \caption{}
  \end{subfigure}
  \begin{subfigure}[b]{0.48\columnwidth}
  \includegraphics{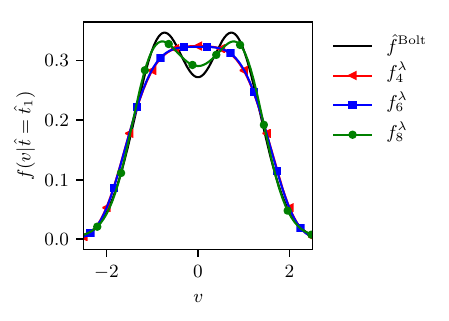}
  \caption{}
  \end{subfigure}
  \begin{subfigure}[b]{0.48\columnwidth}
  \includegraphics{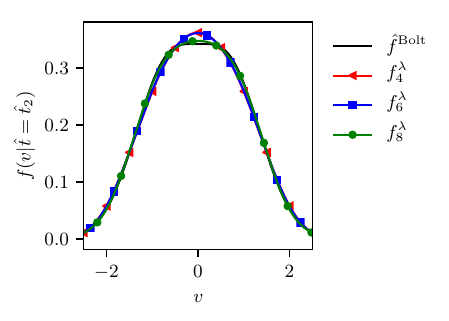} 
  \caption{}
  \end{subfigure}
  \begin{subfigure}[b]{0.48\columnwidth}
  \includegraphics{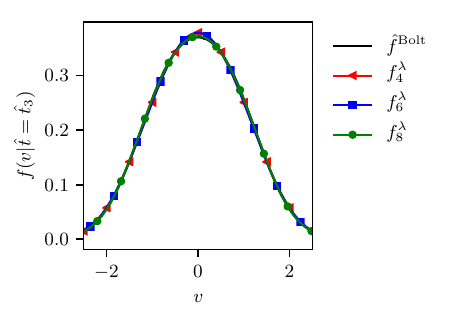}
  \caption{}
  \end{subfigure}
  \caption{Estimating solution of the Boltzmann equation at time $\hat{t}\in \{  5.8, 6.5, 7.5, 8.5\}$ by devised GP-accelerate MED for $N=4,6$ and $8$ moments.}
  \label{fig:Boltz}
\end{figure}

\begin{figure}
\centering
  \begin{subfigure}[b]{0.48\columnwidth}
  \includegraphics{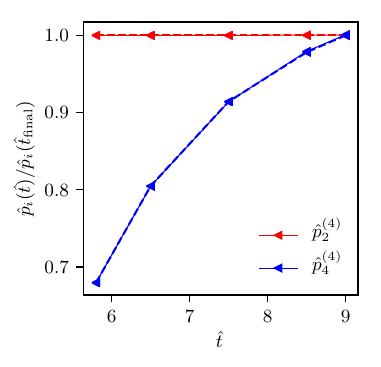}
  \caption{}
  \end{subfigure}
  \begin{subfigure}[b]{0.48\columnwidth}
  \includegraphics{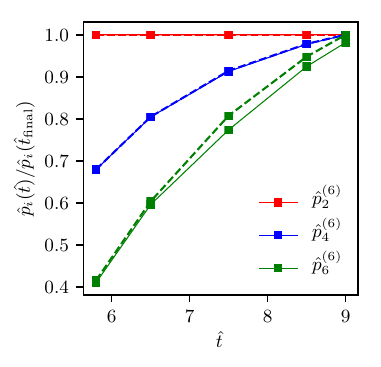}
  \caption{}
  \end{subfigure}
  \begin{subfigure}[b]{0.48\columnwidth}
    \includegraphics{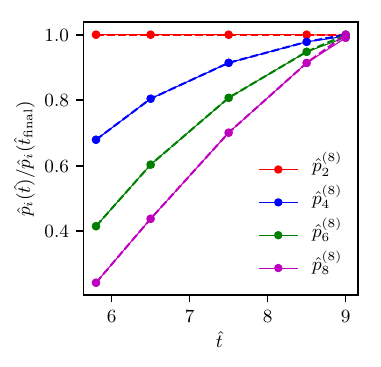}
  \caption{}
  \end{subfigure}
  \begin{subfigure}[b]{0.48\columnwidth}
  \includegraphics{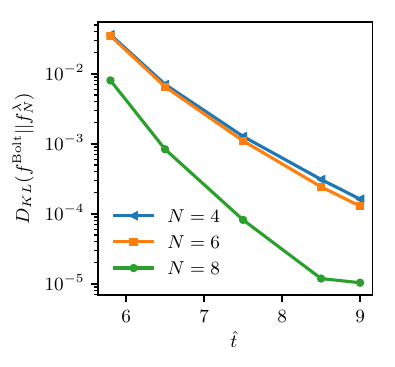}
  \caption{}
  \end{subfigure}
  \caption{Relaxation of standardized moments $\hat{p}^{(N)}\in \mathbb{R}^N$ obtained from exact solution of the Boltzmann equation and devised MED estimation $f^\lambda_N$ with $N\in\{4,6,8 \}$ depicted by dashed and solid lines, respectively, together with the KL divergence between solutions.}
  \label{fig:Boltz_mom_kl}
\end{figure}

\noindent  In this section, we investigate accuracy of the devised data-driven MED in estimating an exact solution of the Boltzmann equation. 
Consider the homogeneous and dimension-less Boltzmann equation in the velocity space $v$ 
\begin{flalign}
\frac{\partial f(v,\hat{t})}{\partial \hat{t}}
=
\frac{1}{4 \pi}
 \int \int [f(v^\prime,\hat{t})f(w^\prime,\hat{t})-f(v,\hat{t})f(w,\hat{t})]
\phi(\mathcal{X}) d\bar{\Omega} dw,
\end{flalign}
\noindent where $\hat{t}$ is the normalized time, superscript $(.)^\prime$ denotes pre-collision velocities of the collision pair, $w$ is the velocity of the collision partner and   $d\bar{\Omega}=\sin(\mathcal{X})d\mathcal{X}d\epsilon_0$ with scattering angle $\mathcal{X}$ and $\epsilon_0\in[0,2\pi]$. Here $g=|v-w|$ is the magnitude of relative velocity.
Note that in the case of the isotropic scattering we have $\phi(\mathcal{X})=1$. As shown in \cite{krook1977exact}, an exact solution 
\begin{flalign}
f^{\mathrm{Bolt}}(v,\hat{t}) = \frac{\exp(-v^2/2K(\hat{t}))}{2 K(\hat{t}) [2\pi K(\hat{t})]^{3/2}} \left[  (5K(\hat{t})-3) + \frac{1-K(\hat{t})}{K(\hat{t})} v^2 \right]/\mathcal{I}
\label{eq:ex_boltz}
\end{flalign}{}
can be obtained, where $\mathcal{I}$ is the normalizing factor and
\begin{flalign}
K(\hat{t}) = 1-\exp(\hat{t}/6)~.
\end{flalign}
\noindent Note that Eq.~\eqref{eq:ex_boltz} provides a valid solution of the Boltzmann equation once $\hat{t}\geq 6 \log(5/2)$.  As derived in \cite{krook1977exact}, the even moments for this isotropic setting evolve according to
\begin{flalign}
p_{2n}(\hat{t}) &= \frac{(4n+1)!}{2^{2n} (2n)!} M_{2n} \ \ \ \text{and}\\
M_{2n}(\hat{t}) &= K^{2n-1}(\hat{t}) \left[ 2n-(2n-1)K(\hat{t}) \right], 
\end{flalign}
for $n\in\{0,1,...\}$. In order to deploy our GP estimator of MED, the input moments need to be standardized via
\begin{flalign}
\hat{p}_{k}(\hat{t}) &= \frac{p_{k}(\hat{t})}{p_{2}(\hat{t})^{k/2}}, \ \ \text{for }k=1,...,N~.
\end{flalign}
By plugging standardized moments at any time $\hat{t}$ as the input in the trained GP, the outcome Lagrange multipliers are predicted.
As shown in  Figs.~\ref{fig:Boltz}-\ref{fig:Boltz_mom_kl}, the trained GP-accelerated MED estimator provides an accurate solution of the Boltzmann equation. As expected, the accuracy in prediction improves once more moments are considered. 

\section{Conclusions}
\label{sec:conclusions}
\noindent The moment closure problem arising from high dimensional systems continues to be a challenge for scientific computing. While MEDs offer an interesting solution framework for estimating the underlying probability density from a given set of moments, the computational cost associated with computing the Lagrange multipliers hindered their use for practical settings. In this study, we accelerate finding the MED by employing GPs as a regression map from moments to Lagrange multipliers.  By taking advantage of the fact that computing the moments from Lagrange multipliers can be performed by simple numerical integrations, around 1000 training data points were generated. Appropriate preparation of the training set by ensuring zero mean and unity variance of MED, besides careful choice of the kernel function have been carried out for a one-dimensional bounded sample space.  The results of capturing bi-modal distributions, noisy distributions, and BGK/Boltzmann type relaxations show encouraging performance of the GP-accelerated MED. However, GP prediction of Lagrange-multipliers becomes less accurate once moments near the realizability limit are encountered. This issue can be tackled by enriching the training points near those limits. For future studies, higher dimensional sample spaces besides sparse GPs will be pursued to further generalize the devised scheme.

\section{Acknowledgment}
\noindent Hossein Gorji acknowledges the funding provided by Swiss National Science Foundation under the grant number~174060. Manuel Torrilhon and Mohsen Sadr acknowledge the funding provided by German Research Foundation (DFG) with the number IRTG-2379~.

\bibliographystyle{elsarticle-num} 
  \bibliography{references}

\end{document}